\documentclass[pmlr]{jmlr}

\RequirePackage{graphicx}
 \usepackage{booktabs}
\usepackage{longtable}
\usepackage{siunitx}

\usepackage[switch]{lineno}

\usepackage{times,amsmath}
\usepackage{bbm}
\usepackage[normalem]{ulem}
\useunder{\uline}{\ul}{}
\usepackage{multirow}

\theorembodyfont{\upshape}
\theoremheaderfont{\scshape}
\theorempostheader{:}
\theoremsep{\newline}

\jmlrvolume{298}
\jmlryear{2025}
\jmlrworkshop{Machine Learning for Healthcare}

\title[LEAVES: Learning Views]{LEAVES: Learning Views for Time-Series Biobehavioral Data in Contrastive Learning}

\author{%
\Name{Han Yu} \Email{han.yu@rice.edu}\\
\addr Rice University, USA
\AND
\Name{Huiyuan Yang} \Email{hyang@mst.edu}\\
\addr Missouri University of Science \& Technology, USA
\AND
\Name{Akane Sano} \Email{akane.sano@rice.edu}\\
\addr Rice University, USA
}

\begin{document}

\maketitle

\begin{abstract}
Contrastive learning has been utilized as a promising self-supervised learning approach to extract meaningful representations from unlabeled data. The majority of these methods take advantage of data-augmentation techniques to create diverse views from the original input. However, optimizing augmentations and their parameters for generating more effective views in contrastive learning frameworks is often resource-intensive and time-consuming. While several strategies have been proposed for automatically generating new views in computer vision \citep{tamkin2020viewmaker, rusak2020increasing}, research in other domains, such as time-series biobehavioral data, remains limited. In this paper, we introduce a simple yet powerful module for automatic view generation in contrastive learning frameworks applied to time-series biobehavioral data, which is essential for modern health care, termed \textbf{lea}rning \textbf{v}i\textbf{e}ws for time-\textbf{s}eries data (LEAVES). This proposed module employs adversarial training to learn augmentation hyperparameters within contrastive learning frameworks. We assess the efficacy of our method on multiple time-series datasets using two well-known contrastive learning frameworks, namely \textit{SimCLR} and \textit{BYOL}. Across four diverse biobehavioral datasets, LEAVES requires only ~20 learnable parameters—dramatically fewer than the ~580,000 parameters demanded by frameworks like ViewMaker, previously proposed adversarially trained convolutional module in contrastive learning, while achieving competitive and often superior performance to existing baseline methods. Crucially, these efficiency gains are obtained without extensive manual hyperparameter tuning, which makes LEAVES particularly suitable for large-scale or real-time healthcare applications that demand both accuracy and practicality. The code of this work is available at: \url{https://github.com/comp-well-org/LEAVES}.
\end{abstract}



\section{Introduction}
\label{sec:intro}
Modern sensing devices collect continuous time-series data from the human body and provide opportunities for monitoring health and behavior. Researchers have demonstrated that such time-series data are crucial for the development of innovative methods for patient monitoring, diagnosis, and treatment \citep{goldberger2000physiobank}. For instance, with machine learning algorithms, electrocardiogram (ECG) has been used in disease diagnosis such as arrhythmia \citep{smigiel2021ecg, li2021bat, zhang2021mlbf} and sleep apnea \citep{fang2022sleep, chang2020sleep, singh2019novel}; electroencephalograph (EEG) has been leveraged for the sleep stage recognition \citep{perslev2019u, mousavi2019sleepeegnet, phan2022sleeptransformer}. 

While the aforementioned methods generally require decent amounts of high-quality annotations to construct reliable and robust machine learning models, acquiring such labels for training deep learning models on biobehavioral data is often challenging. This shortage has encouraged researchers to leverage unsupervised pre-training approaches. Consequently, self-supervised learning techniques, including contrastive learning, have been widely used to improve the robustness of the model \citep{yuan2022self, mehari2022self, sarkar2020self, shah2021evaluating}. For instance, Shah et al. \citep{shah2021evaluating} employ contrastive learning frameworks such as SimCLR \citep{chen2020simple} and BYOL \citep{grill2020bootstrap} to maximize the agreement between the original and augmented electrocardiogram (ECG) samples. This pre-training model outperforms supervised baselines in several downstream health-related tasks, including the detection of sleep apnea, hypertension, and diabetes. Data augmentation is a key element of the contrastive learning methods, which aims to create diverse modifications of the original input for pre-text task. However, selecting effective data augmentation methods for the diverse types of time series biobehavioral data remains an open challenge \citep{yang2022empirical}.

To address the challenge, different approaches have been proposed to find more efficient ways to search for more effective data augmentation methods, such as \citep{tamkin2020viewmaker, hu2021adco, qiu2021neural, rusak2020increasing}.
These methods generate reasonably corrupted views for image datasets that improve model performance. For example, \citep{tamkin2020viewmaker} introduced ViewMaker, an adversarially trained convolutional module in contrastive learning, to generate augmentations for images. However, methods such as ViewMaker may not be suitable for time-series biobehavioral data. First, image-based methods introduce uninterpretable noise to the original signal and result in unfaithful augmented views, where the signal waveform becomes overcorrupted. Second, previous methods primarily focus mainly on distortions to the signal's magnitude, which neglects critical temporal and frequency domain information \citep{um2017data, yang2022empirical, mehari2022self, raghu2022data}.

In this work, we introduce LEAVES (Learning Views for Time-Series Data), a lightweight module that automates augmentation policy tuning for time-series biobehavioral data within contrastive learning frameworks. LEAVES is adversarially trained to generate challenging yet faithful augmentations, which enhances the ability of encoders to learn robust representations. Further, to aid in the extraction of meaningful representation from both the temporal and frequency domains, we introduce two novel differentiable augmentation approaches, Time Modulation \textit{(TimeM)}  and Frequency Suppression \textit{(FreqSup)}, that can provide appropriate and smooth distortions in the temporal and frequency domains respectively. Further, comprehensive experiments demonstrate that LEAVES achieves competitive performance compared to manually fine-tuned augmentation policies within SimCLR and BYOL frameworks while significantly reducing the search cost for optimal augmentations.
Our contributions can be summarized as follows:
\begin{itemize}
    \item We introduce LEAVES, a novel method for automatically learning effective views in contrastive learning frameworks for time-series biobehavioral data. To our knowledge, this is the first study to explore automatic data augmentation for contrastive learning using time-series biobehavioral data.
    \item We propose two differentiable data augmentation methods including \textit{TimeM} and \textit{FreqSup} for adding subtle time-domain and frequency-domain transformation to the data. 
    \item Our comprehensive experiments demonstrate that LEAVES achieves competitive performance compared to manually fine-tuned augmentation policies within SimCLR and BYOL frameworks while significantly reducing the search cost for optimal augmentations.
\end{itemize}

\subsection*{Generalizable Insights about Machine Learning in the Context of Healthcare}
Time-series biobehavioral signals—including ECG, EEG, and Inertial Measurement Unit (IMU) data—are increasingly pivotal in modern, machine-learning-driven healthcare systems. Nevertheless, the scarcity of high-quality labels often impedes fully supervised approaches, which causes a shift toward self-supervised learning for more robust representation learning. In practice, many contrastive learning methods still adopt suboptimal data augmentation policies, primarily because tuning these policies can be both time- and resource-intensive.
LEAVES addresses this challenge by integrating data augmentation policy tuning directly into the neural network and optimizing these policies via adversarial training—all with minimal overhead. This simple and computationally efficient design can be applied to a wide array of time-series biobehavioral signals for various downstream tasks. Moreover, we show that LEAVES consistently achieves competitive or even outperforms manually tuned augmentation policies and state-of-the-art contrastive learning baselines, while introducing only a small number of extra parameters.

\section{Related Work}
\subsection{Augmentation-Based Contrastive Learning}
Contrastive learning algorithms have been leveraged in cutting-edge self-supervised deep learning methods, with data augmentation serving as a core element in generating diverse views from original inputs to create contrastive pairs. A series of contrastive learning frameworks developed for image transformation within computer vision applications have been introduced \citep{he2020momentum, chen2020simple, grill2020bootstrap, chen2021exploring, tamkin2020viewmaker, zbontar2021barlow, wang2022contrastive, zhang2022rethinking}. Among them, SimCLR \citep{chen2020simple} and BYOL \citep{grill2020bootstrap} are the two most widely used frameworks. For instance, SimCLR \citep{chen2020simple} aims to enhance the agreement between two distinct augmented views of a single image, while BYOL \citep{grill2020bootstrap} employs a cooperative learning environment between a target and an online network using two transformed views of an image. Other methods, such as Barlow Twins \citep{zbontar2021barlow}, VICReg \citep{bardes2022vicreg}, MoCo-v2 \citep{chen2020improved}, MoCo-v3 \citep{chen2021empirical}, and SimSiam \citep{chen2021exploring}, have further refined the principles of contrastive and redundancy-reduction learning. Although these newer approaches have shown promise, their adaptation to non-visual modalities, particularly time-series biobehavioral data, remains limited.

\subsection{Contrastive Learning in Time-Series Biobehavioral Data}
The advances in self-supervised learning have inspired the adaptation of contrastive learning techniques to time-series biobehavioral data \citep{yue2022ts2vec, gopal20213kg, mehari2022self, wickstrom2022mixing, eldele2021time, yue2022ts2vec, luo2023time, hallgarten2023ts}. For example, Gopal \textit{et al.} \citep{gopal20213kg} proposed a domain-knowledge-infused augmentation for ECG data, formulating views conducive to contrastive learning. Mehar \textit{et al.} \citep{mehari2022self} extended established methods including SimCLR, BYOL, and CPC \citep{oord2018representation} to time-series ECG data to improve clinical task performance. \cite{hallgarten2023ts} adapted MoCo in EEG data for human activities recognition. Despite these advancements, the empirical selection of data augmentations for view generation is not always optimal, especially for new or less studied datasets, which posts the exploration of augmentation policy as a costly problem.

Although carefully selected augmentations can improve model performance\citep{yue2022ts2vec, mehari2022self}, inappropriate augmentation policies, on the other hand, can negatively impact model performance, particularly with time series biobehavioral data \citep{yang2022empirical}.  A robust approach to selecting data augmentation policies is essential in contrastive learning applications, especially with sensitive time-series biobehavioral data. Nevertheless, to the best of our knowledge, there is no existing study that focuses on augmentation policy exploration for time-series biobehavioral data in contrastive learning.

\subsection{Automatic Augmentation}
Several methods have been proposed to improve augmentation strategies, providing alternatives to traditional empirical approaches \citep{cubuk2019autoaugment, ho2019population, lim2019fast, li2020dada, cubuk2020randaugment, liu2021direct}. AutoAugment \citep{cubuk2019autoaugment}, for instance, employs a reinforcement learning approach to traverse through augmentation policies and optimize the weights and orders of diverse augmentation techniques. DADA \citep{li2020dada} uses a gradient-based optimization approach to identify the most effective augmentation policy during training, significantly reducing training time compared to earlier methods.

In addition to exploring the augmentation policy space, several studies have investigated auto-generating views, where data transformations are generated by neural networks \citep{tian2020makes, rusak2020increasing, tamkin2020viewmaker}. Specifically, \cite{rusak2020increasing} utilized a CNN model to introduce noise into the input data and adversarially optimize the perturbation generator with respect to a supervised loss. Similarly, \cite{tamkin2020viewmaker} implemented a ResNet-based ViewMaker module to generate views for contrastive learning. However, these methods are primarily capable of altering the amplitude of the original signals and are less effective in modulating crucial temporal and frequency domain information in time-series data. In contrast, our proposed LEAVES module not only modulates the original signals in both time and frequency domains but also ensures that the augmented views remain faithful to the underlying biobehavioral patterns—a critical requirement for clinical applications.

\section{Methodology}

\begin{figure*}
\centering
\includegraphics[width=0.8\linewidth]{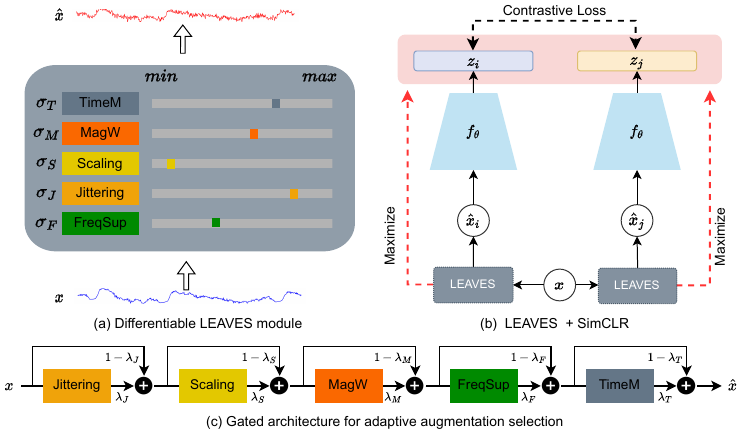}
\vspace{-.2cm}
\caption{Overview of LEAVES integrated within the SimCLR contrastive learning framework. (a) LEAVES generates augmented views ($\hat{x}$) from input $x$ using differentiable transformations with learnable intensity ($\sigma$) parameters. (b) Augmented views are encoded to representations ($z$) via encoder $f_\theta$. The framework is trained adversarially with a contrastive loss. (c) Gating network enables adaptive augmentation selection by activating/deactivating transformations based on learned $\lambda$ values. }
\label{fig:contrastive_framework}
\end{figure*}

In this study, we develop our method on two well-known contrastive learning algorithms, including SimCLR \citep{chen2020simple} and BYOL \citep{grill2020bootstrap}. The overall architecture for the pre-training method demonstrated with SimCLR is illustrated in Figure \ref{fig:contrastive_framework}. Both SimCLR and BYOL utilize 1D ResNet18 as encoders.

In this section, we first introduce a differentiable LEAVES module designed to generate challenging yet faithful views for time-series biobehavioral inputs. Following this, we detail how the LEAVES module is seamlessly integrated into the contrastive learning framework to enable efficient view generation. Then, we describe the adversarial training method to optimize LEAVES with contrastive learning frameworks.

\subsection{LEAVES}
This section introduces LEAVES, a lightweight module designed for seamless integration into existing contrastive learning frameworks. LEAVES generates challenging views while maintaining the original input's essential waveforms. It employs various differentiable data augmentation methods, including \textit{Jittering} ($\mathcal{T}_{J}$), \textit{Scaling} ($\mathcal{T}_{S}$), magnitude warping (\textit{MagW}, $\mathcal{T}_{M}$), time modulation (\textit{TimeM}, $\mathcal{T}_{T}$), and frequency suppression (\textit{FreqSup}, $\mathcal{T}_{F}$). These augmentations are sequentially applied as indicated by the symbol $\odot$. For instance, $\mathcal{T}_{J} \odot \mathcal{T}_{P}$ indicates the application of jittering noise followed by data permutation.

For a given dataset $x \in \mathbb{R}^{C \times L}$, where $L$ represents the length of time series data and $C$ the number of channels, the transformed view $\hat{x}$ is represented as:

\begin{equation}
    \begin{aligned}
    \label{eq:da}
    \mathbf{\hat{x}} = \mathbf{x} & \odot \left( \lambda_J \mathcal{T}_J(\sigma_J) \right) \odot \left( \lambda_S \mathcal{T}_S(\sigma_S) \right) \odot \left( \lambda_M \mathcal{T}_M(\sigma_M) \right) \\
    & \odot \left( \lambda_F \mathcal{T}_F(\sigma_F) \right) \odot \left( \lambda_T \mathcal{T}_T(\sigma_T) \right)
    \end{aligned}
\end{equation}

Here, $\sigma$s are the hyperparameters controlling the intensity of augmentations applied to the original sample with varying ranges across augmentations. For instance, $\sigma$s $\in [0, 0.10]$ specifies the standard deviation for noise generation with \textit{Jittering}, \textit{Scaling}, \textit{MagW}, and \textit{TimeM}; whereas the $\sigma_{F} \in [0, 1]$ is the suppression rate that controls the intensity of \textit{FreqSup}.  LEAVES not only seeks to fine-tune the optimal values of $\sigma$s but also introduces a gated network architecture with gating parameters $\lambda$s that enables adaptive augmentation selection via sigmoid activation. This activation produces values always close to 0 or 1 by introducing the steepness-controlling value into the sigmoid, which enables differentiable "soft" gating. We initially set $\lambda$s to 1, which initially enables all the augmentations initially.  This is achieved by initializing the learnable pre-sigmoid inputs,  denoted as $\sigma_k$ for each gate $k$, to a sufficiently large positive value.  Figure 5 (lower row) illustrates the evolution of these gate statuses, starting from the 'on' state for all augmentations, which corresponds to $\lambda_k$ close to 1. These gates can then be learned as "on" or "off" during training. This enhances LEAVES' ability to autonomously determine the most effective augmentation combinations for any given input.
LEAVES aims to fine-tune the optimal values of $\sigma$s and $\lambda$s, which optimizes both the intensity and presence of each augmentation method. This approach encourages the creation of various views by learning combinations of different augmentation techniques. 
The order of augmentations in Equation \ref{eq:da} is fixed in this study, as the model performance remains robust within the change of order, as indicated in Supplemental Material \ref{ab:aug_order}.

The LEAVES module, along with the representation encoder, is trained in an adversarial manner within the SimCLR framework as depicted in Figure \ref{fig:contrastive_framework}. LEAVES aims to generate views that minimize agreement between representation pairs, whereas the encoder is learning to maximize agreement among different views. This adversarial process encourages the encoder to learn robust representations that capture the underlying signal despite the transformations introduced by LEAVES.

\begin{figure*}
\centering
  \includegraphics[width=.98\linewidth]{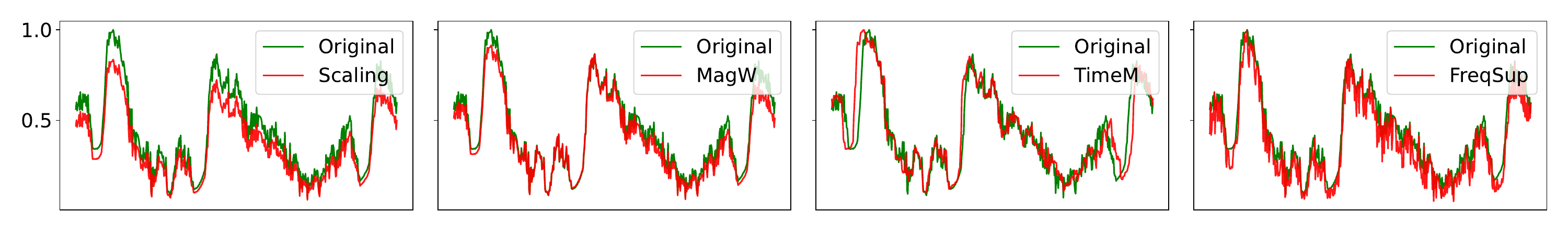}
  \vspace{-.5cm}
  \caption{Examples of different data augmentation methods for time-series biobehavioral data (magnitude $\sigma$ = 0.03).}
  \label{fig:DA}
  \vspace{-.5cm}
\end{figure*}

\begin{figure}
\centering
\includegraphics[width=0.5\linewidth]{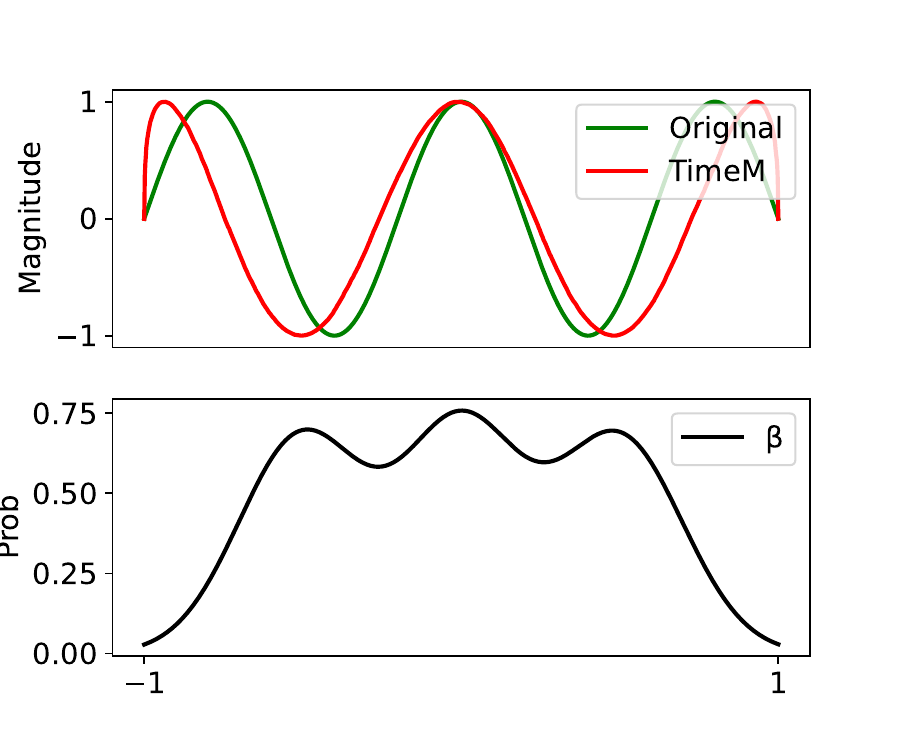}
\vspace{-.5cm}
\caption{An illustration of time modulation augmentation deploying three Gaussian components. Areas of higher probability in the Gaussian mixture models are up-sampled in the transformation, and conversely, those of lower probability are down-sampled.}
\label{fig:time_dis}
\vspace{-.5cm}
\end{figure}

\subsubsection{Differentiable Data Augmentations for Time-Series Data}
LEAVES incorporates a set of data augmentation approaches, with each providing different distortions from the original signal. For example, \textit{Jittering, Scaling}, and \textit{MagW} introduce variations in the magnitude of the original signals; while \textit{TimeM} and \textit{Perm} alter the temporal structure, and \textit{FreqSup} alters information in the frequency domain.

To efficiently tune the hyperparameters of augmentation methods, we design to make these hyperparameters tune within the training process. Nevertheless, a challenge arises from the non-differentiable nature of operations such as random value sampling and indexing in these methods. We address this by applying reparameterization techniques to make these operations differentiable. Further, to introduce the temporal distortion, we introduce the \textit{TimeM} method for introducing temporal distortion, as described in the following section and demonstrated in Figure \ref{fig:DA}. Additionally, we propose a frequency-based augmentation method named frequency suppression (\textit{FreqSup}) to modify the signal by changing frequency domain information. To avoid the generated views being over-corrupted from the original signals, we set constraints on the augmentation methods: a limit of $\eta = 0.10$ on the $\sigma$ values for magnitude-based methods, a maximum value of standard deviation from 0 - 1.0 in each component of GMM \textit{TimeW}, and a minimum $\sigma$ value of 0 in \textit{FreqSup}.

\textbf{Jittering} infuses the original signal with randomly generated noise with a Normal distribution:
\begin{equation}
\label{eq:jitter_revised}
\mathbf{\hat{x}} = \mathbf{x} + \mathbf{\epsilon}_J, \quad \epsilon_J \sim \mathcal{N}(0, \sigma_J^2)
\end{equation}
Here, $\mathbf{\epsilon}_J \in \mathbb{R}^{C \times L}$ represents the noise introduced, with $\sigma_J$ as the adjustable hyperparameter.

\textbf{Scaling} manipulates the signal’s amplitude, entailing multiplication by randomly derived factors specific to each channel:
\begin{equation}
\label{eq:scaling_revised}
\mathbf{\hat{x}} = [\hat{x}_1, \hat{x}_2, ... ,\hat{x}_C] = [\epsilon_S^1 x_1, \epsilon_S^2 x_2, ... \\
,\epsilon_S^C x_C]
\end{equation}
$\mathbf{\epsilon}_S \sim \mathcal{N}(1, \sigma_S^2) \in \mathbb{R}^{C}$ denotes the sampled factors, with $\sigma_S$ as the essential hyperparameter.

\textbf{MagW (Magnitude Warping)} \citep{um2017data} involves distorting the magnitude of the original signal with a randomly generated smooth curve. First, we sample $k$ nodes from $\mathcal{N}(1, \sigma_M^2)$, which yields $knot \in \mathbb{R}^{ C \times k}$. Then, we interpolate $knot$s evenly with a linear function producing $\epsilon_{M}$. The modified view can be denoted as:

\begin{equation}
\label{eq:magw}
    \mathbf{\hat{x}} = \mathbf{x} + \mathbf{\epsilon}_M
\end{equation}
We keep $k$ to 8 and concentrate on the hyperparameter $\sigma_M$ in this study.

\textbf{TimeM} modifies the temporal location within the original sequences by generating probabilities to determine which locations in the original signals should be sampled. By using a reparameterized Gaussian mixture model with $M$ components, represented as $\sum_i^M \phi_{i}\mathcal{N}(\mu_i, \sigma_i^2)$,  \textit{TimeM} generates probabilities $\beta \in \mathbb{R}^{ C \times L}$ ranging from 0 to 1 to warp the temporal positions. 
$\mu$ controls the center of each warping region, and $\sigma$ defines its width, which affects the intensity of time warping around $\mu$. Regions with higher $\beta$ are upsampled, while those with lower $\beta$ are downsampled. An example of \textit{TimeM} is shown in Figure \ref{fig:time_dis}, -1 corresponds to the first time step (position 1) in the original signal, while 1 is the last time step (position $L$). 

In this study, we empirically fix $M = 7$ and set the $\mu$ values at [-0.75, -0.5, -0.25, 0, 0.25, 0.5, 0.75]. This setting is based on preliminary hyperparameter tuning that suggests that evenly distributed seven components between -1 to 1, which offers a balanced trade-off between model complexity and the ability to capture variations in the signal. On the other hand, we make the standard deviation $\sigma$ a target parameter for LEAVES to learn.

\textbf{FreqSup} is a frequency-based augmentation method that modifies the signal by manipulating its frequency components. This method employs a soft mask to selectively suppress certain frequencies. The mask is designed to preserve frequencies bands with high amplitudes to prevent excessive distortion of the original signal. First, a suppression mask ($MASK_{S}$) is defined to introduce the probability of suppression.

\begin{equation}
\text{MASK}_S = sigmoid\left( (U(0,1) - \sigma_F) \right)
\end{equation}
where $U(0,1)$ is a uniform random variable between 0 and 1, $\sigma_F$ is a learnable parameter that controls the suppression rate. In addition, we introduce a projection mask ($MASK_P$) to preserve high-amplitude frequency components.

\begin{equation}
\text{MASK}_P = 0.5 \cdot \left(\tanh\left(\frac{\text{Amp}}{\max(\text{Amp})} \right) + 1 \right)
\end{equation}
where `Amp` represents the amplitude of the frequency spectrum of the signal

\begin{equation}
\text{MASK}_F = (1 - \text{MASK}_P) \cdot \text{MASK}_S
\end{equation}
$\text{MASK}_F$ is the combination of $\text{MASK}_S$ and $\text{MASK}_P$.

\begin{equation}
\mathbf{\hat{x}} = \text{IFFT}\left( \text{FFT}(\mathbf{x}) \cdot \text{MASK}_F \right)\\
\end{equation}
FFT and IFFT denote the Fast Fourier Transform and its inverse, respectively.

Thus, the \textit{FreqSup} method corrupts the signal by suppressing low-amplitude components in the frequency domain, which are determined by $\sigma_F$. Smaller values of $\sigma_F$ lead to a boarder range of affected frequency bands, which then results in a stronger augmentation.

\begin{figure}
\centering
\includegraphics[width=0.5\linewidth]{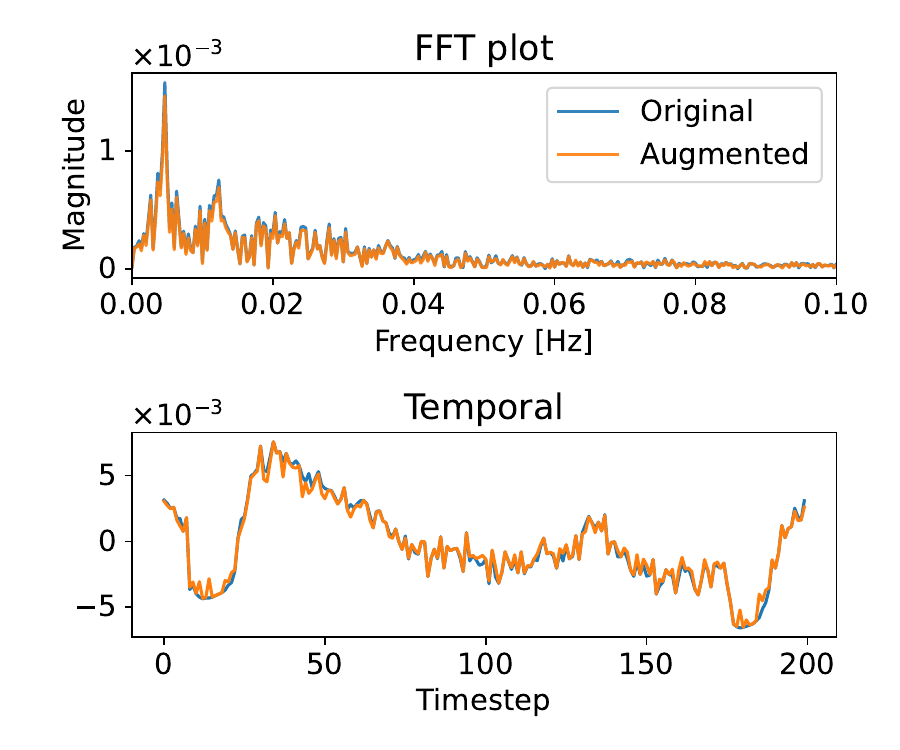}
\vspace{-.5cm}
\caption{The upper plot shows the original signal and the augmented signal in the frequency domain. The augmentation selectively reduces the magnitude of specific frequency components. The impact of frequency suppression in the time domain is shown in the lower plot.}
\label{fig:freqsuppress}
\vspace{-.5cm}
\end{figure}

\subsection{Adversarial Training Approach}
To optimize both the LEAVES augmentation module and the encoder, we employ an adversarial training strategy. We use SimCLR as an example to illustrate this approach. We define the output of the encoder in representation learning as $z$. Within the framework, we consider $N$ pairs of representations, denoted as $(z_i,z_j)$, $\{i, j\} \in [1, N]$. The objective is to maximize the agreement between these pairs of views, and the corresponding loss function is defined as:

\begin{equation}
    \mathcal{L} = \frac{1}{2N} \sum_{k=1}^{N}[\ell(2k-1,2k)+\ell(2k,2k-1)],
\end{equation}
\begin{equation}
    \ell_{i,j} = -\log \frac{\exp(s(z_i, z_j)/\tau)}{\sum_{k=1}^{2N}\mathbbm{1}_{k\neq i}\exp(s(z_i, z_k)/\tau)}
\end{equation}

The cosine similarity between $z_i$ and $z_j$ is represented by $s(z_i, z_j)$, and $\mathbbm{1}_{k\neq i}$ is an indicator function that is equal to 1 when $k \neq i$. The temperature parameter $\tau$ is set to 0.05 in this study. As illustrated in Figure \ref{fig:contrastive_framework}, the LEAVES and encoder are optimized in contrary directions. The encoder attempts to minimize $\mathcal{L}$, while LEAVES attempts to maximize it.  During training, the gradients for the encoder parameters ($\theta_E$) are computed to minimize the contrastive loss $L$ (Equation 9), while the gradients for the LEAVES module parameters ($\theta_L$, which include $\sigma$s and the pre-sigmoid inputs $\alpha_k$ for $\lambda_k$s) are computed to maximize the same loss $L$. Maximizing $L$ for LEAVES is equivalent to minimizing $-L$ from the perspective of the LEAVES module's parameters.
This adversarial process forces the encoder to learn robust representations that capture the underlying signal while being robust to the diverse transformations introduced by LEAVES. After training the SimCLR framework, the learned model weights in the encoder structure are used to initialize the model weights for supervised learning in downstream tasks. The LEAVES framework can also be integrated into other frameworks (e.g., BYOL) for generating effective augmented views.

\section{Evaluation} \label{evaluation}
In this section, we evaluate the effectiveness of our proposed approach using various public time-series biobehavioral datasets. These include Apnea-ECG \citep{penzel2000apnea} for sleep apnea detection, Sleep-EDFE \citep{kemp2018sleep} for sleep stages classification, PTB-XL \citep{wagner2020ptb} for arrhythmias detection, and PAMAP2 \citep{reiss2012introducing} for human activity recognition. 

To benchmark our approach, we plug LEAVES into two well-known contrastive learning frameworks, SimCLR \citep{chen2020simple} and BYOL \citep{grill2020bootstrap}, and compare its performance to that of other popular time-series self-supervised approaches.
Also, a supervised 1D ResNet-18 model serves as a reference for the performance achievable without pre-training. The following sub-sections introduce the experimental settings and the results of application domains. 

\subsection{Experimental Settings}
\textbf{Datasets}: We evaluate LEAVES on four diverse public biobehavioral datasets with each addressing distinct tasks.
\begin{itemize}
    \item \textbf{Apnea-ECG (Sleep Apnea Detection) \citep{penzel2000apnea}:} We utilize the Apnea-ECG dataset to explore the correlation between sleep apnea diagnosis and cardiac activities captured by ECG signals, which are accessible through Physionet \citep{goldberger2000physiobank}. We follow the same settings as the original data \citep{penzel2000apnea}, which utilize a 100Hz ECG one-minute scale to detect the occurrence of apnea. The training set comprises 17233 samples and the testing set had 17010 samples.
    \item \textbf{Sleep-EDFE (Sleep Stage Classification) \citep{kemp2018sleep}:} Electroencephalography (EEG) plays a crucial role in monitoring human brain activities. To evaluate our methods, we conduct evaluations using the multimodal Sleep-EDF (expanded) dataset, which comprises whole-night sleep recordings of 100Hz Fpz-Cz EEG and electrooculography (EOG) signals. Following \citep{supratak2020tinysleepnet}, we extract 42308 30-second samples that are annotated in 5 sleep stages. During the pre-training, we apply two LEAVES modules for EEG and EOG individually to optimize the corresponding views for each modality. Although 10 or 20-fold cross-validations with various settings are commonly used in most prior studies \citep{perslev2019u, mousavi2019sleepeegnet, perslev2021u, phan2022sleeptransformer}, we divide the validation sets based on the subject IDs to avoid information leakage during training.
    \item \textbf{PTB-XL (Arrhythmia Diagnosis) \citep{wagner2020ptb}}: Cardiac arrhythmias are a major contributor to the prevalence of cardiovascular diseases, thus necessitating the development of precise and dependable detection methods for clinical use. We employ the PTB-XL dataset \citep{wagner2020ptb}, which consists of 21,837 12-lead, 10-second ECG recordings at a 100 Hz frequency, is divided into five categories of arrhythmias. We follow the split of train, evaluation, and test guidelines outlined in the original publication \cite{wagner2020ptb}, and report the results on the test set. 
    \item \textbf{PAMAP2 (Human Activity Recognition) \citep{reiss2012introducing}:} The PAMAP2 research demonstrates the ability to detect human activities through data collected from biobehavioral sensors. This dataset includes 3 inertial measurement units (IMU) and a heart rate monitor, which are sampled at 100 Hz and upsampled heart rate data. 12 of the 18 physical activities are used in the experiments, as suggested by \citep{moya2018convolutional, tamkin2020viewmaker}.
\end{itemize}

\noindent \textbf{Baselines:}
We benchmark LEAVES against a range of established approaches to comprehensively assess its effectiveness in representation learning for time-series biobehavioral data:
\begin{itemize}
    \item \textbf{Supervised ResNet-18 \citep{he2016deep}:} We adapt the supervised 1D ResNet-18 architecture as a reference for the performance achievable without pre-training. The encoder output is passed to a 2‑layer MLP: 1025, 512, and $C$, where $C$ is the number of task classes.  Each linear layer is followed by BatchNorm, ReLU, and Dropout rate of 0.2. Notably, we also use the ResNet-18 architecture as the backbone for our contrastive learning models.
    \item \textbf{Contrastive Learning Models:}  We compare LEAVES with SimCLR \citep{chen2020simple}, BYOL \citep{grill2020bootstrap}, MoCo-v3 \citep{chen2021empirical}, and VICReg \citep{bardes2022vicreg}. These methods learn data representations by contrasting similar and dissimilar augmented views of the same sample. To ensure a fair comparison, these methods are implemented based on the same augmentation set as the proposed LEAVES. We further explore the challenges of hyperparameter tuning of augmentations in contrastive learning by exploring the $\sigma$s of specific augmentation techniques within SimCLR and BYOL.
    \item \textbf{Other SSL methods:} We include TS2Vec \citep{yue2022ts2vec} and TS-TCC \citep{emadeldeen2022catcc} in our evaluation. These methods are specifically designed for time-series data and achieve SOTA performances in corresponding domains.
\end{itemize}

\subsection{Results}
Table \ref{tab:res} summarizes the performance in terms of macro F1-score (F1) and Area Under the Receiver Operating Characteristic curve (AUROC) , with red bold indicating the best performance per dataset and metric, and blue underline highlights the second-best. The proposed LEAVES method is compared with BYOL, SimCLR, other time-series SSL methods (TS2Vec, TS-TCC), and a supervised ResNet-18-1D model.

Notably, LEAVES-enhanced models outperform or are highly competitive with all other methods. On the Apnea-ECG, Sleep-EDFE, and PTB-XL datasets, LEAVES (BYOL) achieves the highest AUROC. On the PAMAP2 dataset, LEAVES (SimCLR) secures the top performance in both F1-score and AUROC. This demonstrates the effectiveness of LEAVES in learning robust representations for diverse biobehavioral signals and downstream tasks.

To validate the significance of the observed improvements, we conducted independent t-tests comparing the AUROC of both LEAVES(SimCLR) and LEAVES(BYOL) against key baselines for each dataset. The analysis confirms the robust performance of our method. On the Apnea-ECG, Sleep-EDFE, and PTB-XL datasets, both LEAVES variants demonstrated statistically significant improvements over the Supervised baseline and their corresponding best manually-tuned model. For the PAMAP2 dataset, LEAVES(BYOL) also achieved a statistically significant improvement over the Supervised and best manually-tuned BYOL baselines ($p < .05$), though its advantage over TS2Vec was not statistically significant. These results confirm that the performance gains from the LEAVES framework are not only numerically superior but also statistically meaningful across diverse biobehavioral signals and tasks.

Our experiments highlight the practical challenges of manual hyperparameter tuning for data augmentations. As shown by the grayed out entries in Table 1, empirically searching for the optimal augmentation strength ($\sigma$) for SimCLR and BYOL often leads to suboptimal outcomes. This underscores the difficulty of tuning augmentations to diverse time-series datasets and tasks. In contrast, LEAVES automates this process, learning suitable augmentation policies without requiring extensive hyperparameter searches, and and therefore simplifies the application of contrastive learning while delivering superior performance.

\begin{table*}[]
\centering
\scriptsize
\caption{Performance comparison using macro F1-score and AUROC. Results are shown as mean $ \pm $ standard deviation. The \textbf{best} result in each column is bolded. Statistical tests for the AUROC of LEAVES variants are denoted by: $^*$vs. Supervised, $^\dagger$vs. best manually-tuned corresponding baseline (e.g., LEAVES(SimCLR) vs. best SimCLR$_\sigma$). The tuning results of SimCLR and BYOL are also included in this table, with sub-optimal results \textcolor{gray}{grayed out}.}
\vspace{-.3cm}
\label{tab:res}
\begin{tabular}{l|cc|cc|cc|cc}
\hline
                        & \multicolumn{2}{c|}{Apnea-ECG} & \multicolumn{2}{c|}{Sleep-EDFE} & \multicolumn{2}{c|}{PTB-XL} & \multicolumn{2}{c}{PAMAP2} \\ 
Methods                 & F1            & AUROC         & F1            & AUROC         & F1            & AUROC         & F1            & AUROC         \\ \hline \hline
Supervised              & $.755_{\pm .008}$ & $.793_{\pm .010}$ & $.773_{\pm .007}$ & $.895_{\pm .005}$ & $.665_{\pm .012}$ & $.858_{\pm .010}$ & $.908_{\pm .005}$ & $.941_{\pm .004}$ \\
ViewMaker               & $.760_{\pm .011}$ & $.801_{\pm .013}$ & $.770_{\pm .008}$ & $.888_{\pm .006}$ & $.670_{\pm .011}$ & $.864_{\pm .011}$ & $.913_{\pm .006}$ & $.944_{\pm .005}$ \\
TS2Vec                  & $.765_{\pm .008}$ & $.806_{\pm .010}$ & $.777_{\pm .006}$ & $.902_{\pm .005}$ & $.669_{\pm .012}$ & $.867_{\pm .010}$ & \textbf{.928$_{\pm .005}$} & $.955_{\pm .004}$ \\
TS-TCC                  & $.749_{\pm .011}$ & $.788_{\pm .009}$ & $.762_{\pm .009}$ & $.885_{\pm .008}$ & $.672_{\pm .010}$ & $.870_{\pm .009}$ & $.921_{\pm .007}$ & $.944_{\pm .005}$ \\
VICReg                  & $.758_{\pm .010}$ & $.800_{\pm .007}$ & $.782_{\pm .005}$ & $.906_{\pm .006}$ & $.668_{\pm .013}$ & $.865_{\pm .011}$ & $.915_{\pm .006}$ & $.947_{\pm .006}$ \\
MoCo-v3                 & $.763_{\pm .009}$ & $.808_{\pm .009}$ & $.774_{\pm .006}$ & $.898_{\pm .004}$ & $.665_{\pm .012}$ & $.860_{\pm .014}$ & $.911_{\pm .005}$ & $.942_{\pm .004}$ \\
SimCLR$_{\sigma=.01}$ & \textcolor{lightgray}{$.758_{\pm .009}$} & \textcolor{lightgray}{$.795_{\pm .010}$} & \textcolor{lightgray}{$.776_{\pm .006}$} & \textcolor{lightgray}{$.901_{\pm .005}$} & \textcolor{lightgray}{$.667_{\pm .010}$} & \textcolor{lightgray}{$.861_{\pm .012}$} & \textcolor{lightgray}{$.906_{\pm .004}$} & \textcolor{lightgray}{$.938_{\pm .006}$} \\
SimCLR$_{\sigma=.02}$ & \textcolor{lightgray}{$.762_{\pm .008}$} & \textcolor{lightgray}{$.804_{\pm .011}$} & \textcolor{lightgray}{$.780_{\pm .007}$} & \textcolor{lightgray}{$.908_{\pm .007}$} & \textcolor{lightgray}{$.672_{\pm .009}$} & \textcolor{lightgray}{$.869_{\pm .011}$} & \textcolor{lightgray}{$.913_{\pm .005}$} & \textcolor{lightgray}{$.946_{\pm .005}$} \\
SimCLR$_{\sigma=.03}$ & $.768_{\pm .010}$ & $.811_{\pm .010}$ & \textcolor{lightgray}{$.783_{\pm .006}$} & \textcolor{lightgray}{$.910_{\pm .007}$} & $.672_{\pm .011}$ & $.871_{\pm .012}$ & \textcolor{lightgray}{$.916_{\pm .006}$} & \textcolor{lightgray}{$.948_{\pm .005}$} \\
SimCLR$_{\sigma=.04}$ & \textcolor{lightgray}{$.767_{\pm .010}$} & \textcolor{lightgray}{$.809_{\pm .009}$} & \textcolor{lightgray}{$.781_{\pm .007}$} & \textcolor{lightgray}{$.910_{\pm .008}$} & \textcolor{lightgray}{$.670_{\pm .013}$} & \textcolor{lightgray}{$.868_{\pm .012}$} & $.918_{\pm .006}$ & $.951_{\pm .004}$ \\
SimCLR$_{\sigma=.05}$ & \textcolor{lightgray}{$.767_{\pm .008}$} & \textcolor{lightgray}{$.810_{\pm .008}$} & $.784_{\pm .006}$ & $.908_{\pm .007}$ & \textcolor{lightgray}{$.669_{\pm .011}$} & \textcolor{lightgray}{$.866_{\pm .013}$} & \textcolor{lightgray}{$.915_{\pm .007}$} & \textcolor{lightgray}{$.950_{\pm .005}$} \\
BYOL$_{\sigma=.01}$   & \textcolor{lightgray}{$.766_{\pm .007}$} & \textcolor{lightgray}{$.808_{\pm .009}$} & \textcolor{lightgray}{$.774_{\pm .008}$} & \textcolor{lightgray}{$.897_{\pm .009}$} & \textcolor{lightgray}{$.666_{\pm .010}$} & \textcolor{lightgray}{$.864_{\pm .011}$} & \textcolor{lightgray}{$.917_{\pm .004}$} & \textcolor{lightgray}{$.950_{\pm .004}$} \\
BYOL$_{\sigma=.02}$   & \textcolor{lightgray}{$.771_{\pm .010}$} & \textcolor{lightgray}{$.814_{\pm .007}$} & \textcolor{lightgray}{$.777_{\pm .006}$} & \textcolor{lightgray}{$.903_{\pm .007}$} & \textcolor{lightgray}{$.670_{\pm .009}$} & \textcolor{lightgray}{$.868_{\pm .010}$} & \textcolor{lightgray}{$.921_{\pm .004}$} & \textcolor{lightgray}{$.952_{\pm .005}$} \\
BYOL$_{\sigma=.03}$   & \textcolor{lightgray}{$.775_{\pm .009}$} & \textcolor{lightgray}{$.818_{\pm .008}$} & \textcolor{lightgray}{$.779_{\pm .006}$} & \textcolor{lightgray}{$.906_{\pm .006}$} & \textcolor{lightgray}{$.671_{\pm .010}$} & \textcolor{lightgray}{$.869_{\pm .010}$} & $.924_{\pm .005}$ & $.955_{\pm .004}$ \\
BYOL$_{\sigma=.04}$   & $.777_{\pm .009}$ & $.820_{\pm .007}$ & $.780_{\pm .005}$ & $.905_{\pm .005}$ & $.674_{\pm .011}$ & $.873_{\pm .010}$ & \textcolor{lightgray}{$.922_{\pm .006}$} & \textcolor{lightgray}{$.954_{\pm .005}$} \\
BYOL$_{\sigma=.05}$   & \textcolor{lightgray}{$.776_{\pm .011}$} & \textcolor{lightgray}{$.819_{\pm .009}$} & \textcolor{lightgray}{$.776_{\pm .010}$} & \textcolor{lightgray}{$.902_{\pm .009}$} & \textcolor{lightgray}{$.671_{\pm .008}$} & \textcolor{lightgray}{$.871_{\pm .009}$} & \textcolor{lightgray}{$.923_{\pm .006}$} & \textcolor{lightgray}{$.954_{\pm .004}$} \\ \hline
LEAVES$_{(\textit{SimCLR})}$ & $.775_{\pm .009}^{*}$ & $.822_{\pm .008}^{*\dagger}$ & \textbf{.790}$_{\pm .007}^{*\dagger}$ & \textbf{.914}$_{\pm .006}^{*\dagger}$ & $.677_{\pm .009}^{*}$ & $.875_{\pm .008}^{*\dagger}$ & $.921_{\pm .005}^{*}$ & $.950_{\pm .005}$ \\
LEAVES$_{(\textit{BYOL})}$   & \textbf{.784}$_{\pm .008}^{*\dagger}$ & \textbf{.830}$_{\pm .006}^{*\dagger}$ & $.787_{\pm .005}^{*\dagger}$ & $.912_{\pm .005}^{*\dagger}$ & \textbf{.680}$_{\pm .011}^{*\dagger}$ & \textbf{.877}$_{\pm .011}^{*\dagger}$ & $.926_{\pm .003}^{*}$ & \textbf{.957}$_{\pm .004}^{*\dagger}$ \\ \hline
\end{tabular}
\vspace{-.3cm}
\end{table*}

\subsection{Learning Hyperparameters for Augmentations}

\begin{figure}
\centering
  \includegraphics[width=0.8\linewidth]{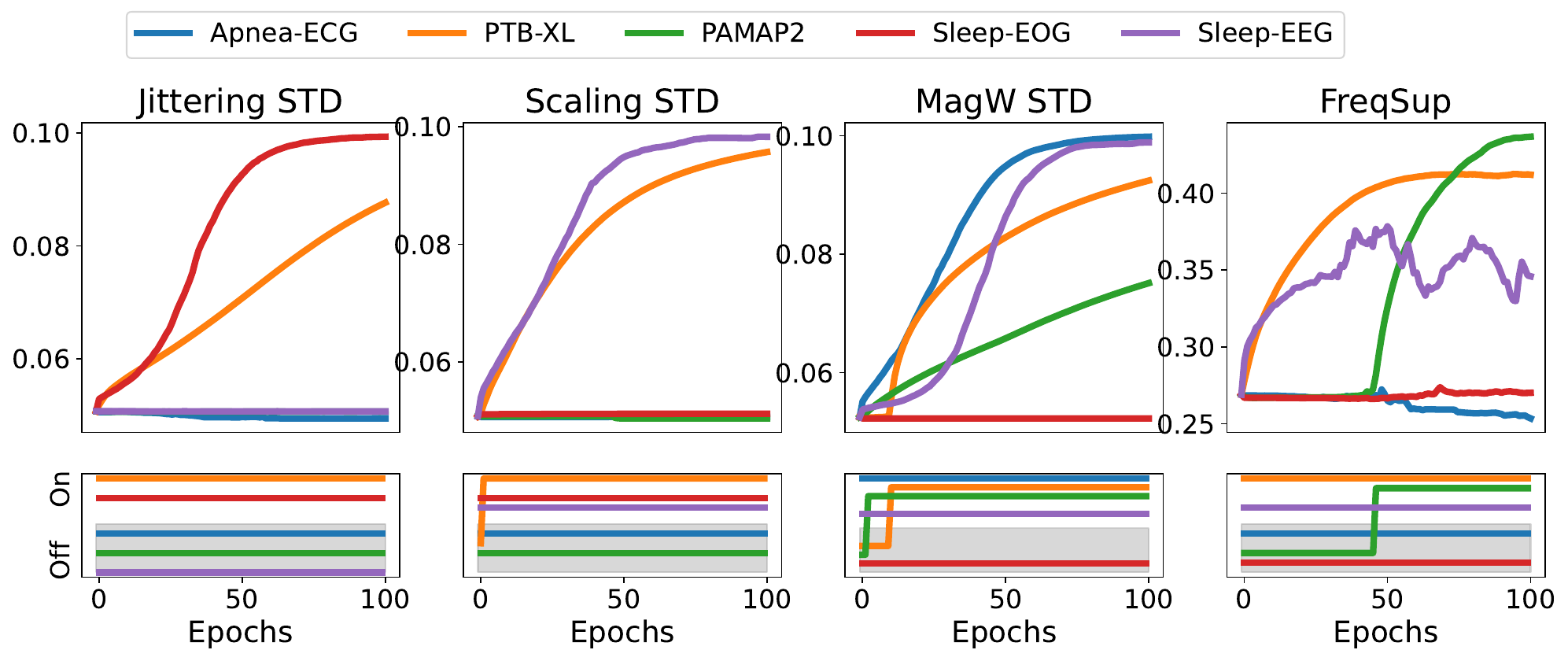}
  \vspace{-.4cm}
  \caption{Visualization of the scalar hyperparameters, including intensity ($\sigma$) and gating status ($\lambda$) during LEAVES (SimCLR) training. The upper row indicates the intensity change of each augmentation; whereas the lower row shows the gating "on" or "off" status of augmentations.
}
\vspace{-0.8cm}
  \label{fig:view_hyperparams}
\end{figure}

In the LEAVES framework, the scalar hyperparameters that control differentiable augmentations are dynamically optimized across training epochs. This adaptability enables LEAVES to adjust augmentation strategies to the unique characteristics and challenges of each dataset. The dynamics of these hyperparameters, recorded at the end of each training epoch, are shown in Figure \ref{fig:view_hyperparams}. We observe the changes in intensity with the on status of the gating parameter.

For \textit{Jittering}, \textit{Scaling}, \textit{MagW}, and \textit{FreqSup}, the $sigma$ values exhibit an increasing trend, which indicates that the augmentations become progressively more aggressive. However, these values do not converge to a single threshold across datasets; rather, they vary across datasets, which reflects the adaptive nature of the framework to different signals. Similarly, the selection of augmentations based on $\lambda$ values shows LEAVES's ability to adapt various augmentation policies to different signals during training. 

The learned parameters of augmentations may reflect the various characteristics of different biobehavioral signals. For Apnea-ECG, \textit{MagW} and \textit{TimeM} are consistently activated, which suggests that altering temporal dynamics and magnitude relationships in the ECG waveform is more informative for 1-lead ECG. In contrast, the 12-lead ECG data in PTB-XL benefits from all augmentations including jittering and scaling that are not activated for Apnea ECG, potentially due to the higher dimensionality and complexity of the 12-lead signal. For PAMAP2 (IMU), the activation of \textit{MagW}, \textit{TimeM}, and \textit{FreqSup} indicate that both temporal and frequency information are essential in learning representation from IMU signals. The initial deactivation of \textit{FreqSup} suggests that LEAVES prioritizes learning temporal patterns before gradually incorporating spectral augmentation into the representations.
\textit{Jittering} and \textit{Scaling} are not activated while pretraining on IMU data. This is potentially because jittering and scaling noise are already present in the IMU data, even without augmentation \citep{nirmal2016noise}.
For Sleep-EDFE, only jittering is activated for EOG while being consistently deactivated for EEG. This suggests that EEG signals are more sensitive to noise perturbations compared to EOG signals. These observations highlight the importance of selecting the proper augmentations based on the specific waveform patterns of the data and task.

\vspace{-.3cm}
\section{Discussion}
In this section, various ablation studies are presented. Additionally, we discuss the learned augmentation policy and computational complexity of the proposed LEAVES module.

\subsection{Ablation Study: Impact of Different Types of Data Augmentations on LEAVES Performances}
In this study, we investigate the impact of the data augmentation methods. We consider the categories of augmentation methods as magnitude-based method (\textit{Mag}), which encompasses adjustments to signal amplitude; temporal method (\textit{TimeM}), which manipulates the temporal positions of signals; and the frequency-based method (\textit{FreqSup}), which modifies the signal from the frequency domain.
We perform an ablation experiment using SimCLR-based LEAVES by systematically removing each category of augmentation method and evaluating the resulting model's performance on various downstream tasks. The results are summarized in Table \ref{tab:ablation_augs}, which showcases the performance drop observed when excluding a specific augmentation method from LEAVES. The table presents results for all four biobehavioral datasets using the drops in macro-F1 score as the performance metric. The results consistently demonstrate that removing any single augmentation technique leads to a decline in performance across most datasets. This indicates that each augmentation method plays a valuable role in enhancing the model's ability to learn informative representations. Notably, both the proposed \textit{FreqSup} and \textit{TimeM} methods consistently contribute to improving performance. Furthermore, the combined effect of removing multiple augmentations often results in more substantial performance drops compared to removing single ones. This suggests that the combinations of different types of augmentations provide more comprehensive and diverse data transformations. 

\begin{table*}[]
    \centering
    \small
  \caption{Performance drops of removing specific augmentation methods from LEAVES. The magnitude-based methods (\textit{Mag}) combine the augmentations of \textit{Jittering}, \textit{Scaling}, and \textit{MagW}. The results are based on the Macro-F1 scores (\%) from the main experimental results.}
  \vspace{-.2cm}
  \begin{tabular}{l|cccccc}
\hline
Dataset    & \textit{Mag (M)}     & \textit{TimeM (T)}   & \textit{FreqSup (F)} & \textit{M \& T} & \textit{M \& F} & \textit{T \& F} \\ \hline\hline
Apnea-ECG  & 2.1 $\downarrow$ & 1.3 $\downarrow$ & 0.1 $\downarrow$ & 2.8 $\downarrow$     & 2.4 $\downarrow$       & 1.3 $\downarrow$         \\
Sleep-EDFE & 1.5 $\downarrow$ & 1.7 $\downarrow$ & 0.8 $\downarrow$ & 2.4 $\downarrow$     & 2.0 $\downarrow$       & 2.0 $\downarrow$         \\
PTB-XL     & 1.4 $\downarrow$ & 0.3 $\downarrow$ & 0.6 $\downarrow$ & 1.5 $\downarrow$     & 1.6 $\downarrow$       & 0.7 $\downarrow$         \\
PAMAP2     & 1.2 $\downarrow$ & 2.2 $\downarrow$ & 1.5 $\downarrow$ & 3.0 $\downarrow$     & 1.9 $\downarrow$       & 2.3 $\downarrow$         \\ \hline
\end{tabular}
\vspace{-.3cm}
\label{tab:ablation_augs}
\end{table*}

\subsection{Comparison of LEAVES with ViewMaker for View Generation}
\begin{figure}
\centering
\includegraphics[width=1\linewidth]{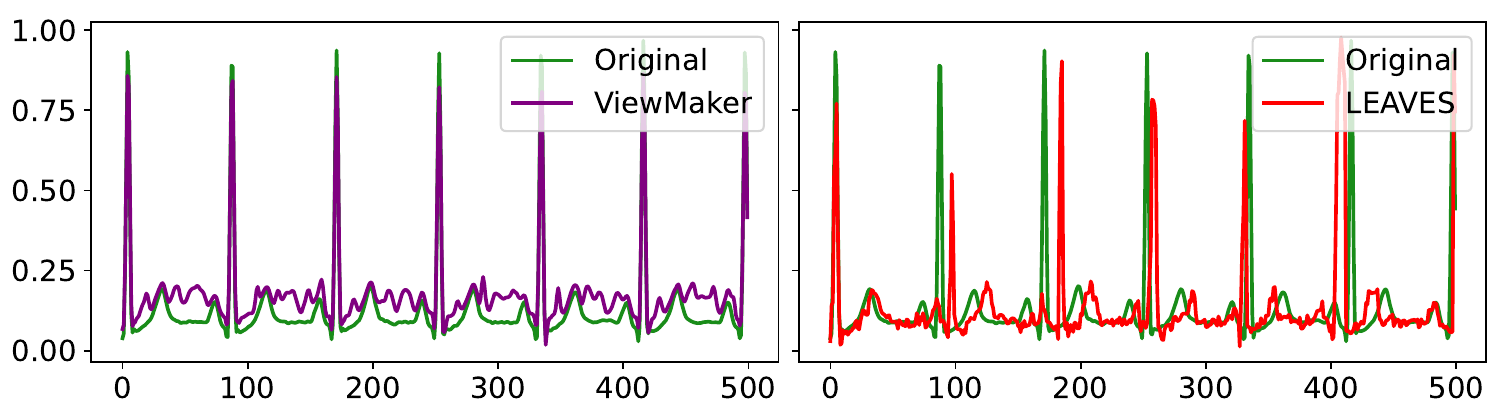}
\caption{Visualization of the views learned using the ViewMaker and the proposed method}
\label{fig:viewmaker_vs_ours}
\end{figure}

Our approach is inspired by ViewMaker \citep{tamkin2020viewmaker}, but there are some key differences. Instead of using "black-box" deep networks to create spatial distortions, our LEAVES module uses time-series domain knowledge-based augmentations to generate views. This makes LEAVES lightweight and can produce more diverse yet reliable views than ViewMaker. Figure \ref{fig:viewmaker_vs_ours} shows an example of ViewMaker's limitations in temporal distortion and information preservation, as ViewMaker distorts most ECG fiducials. To assess the quality of the data from the generated views, we use an ECG quality check method \citep{zhao2018sqi} with the NeuroKit package \citep{Makowski2021neurokit}. We find that ViewMaker corrupted almost half (49.5\%) of the ECGs to be labeled as "Unacceptable," which indicates signals that are barely recognizable as ECG signals. In contrast, the views generated using our proposed method with both SimCLR and BYOL have acceptable data quality (96.0\% and 93.1\%, respectively) compared to the original data. These limitations of ViewMaker when applied to time-series data motivate us to develop our proposed method in this study.

\subsection{Adaptive Learning and Augmentation Strategies in LEAVES}
\begin{figure}
\centering
  \includegraphics[width=.6\linewidth]{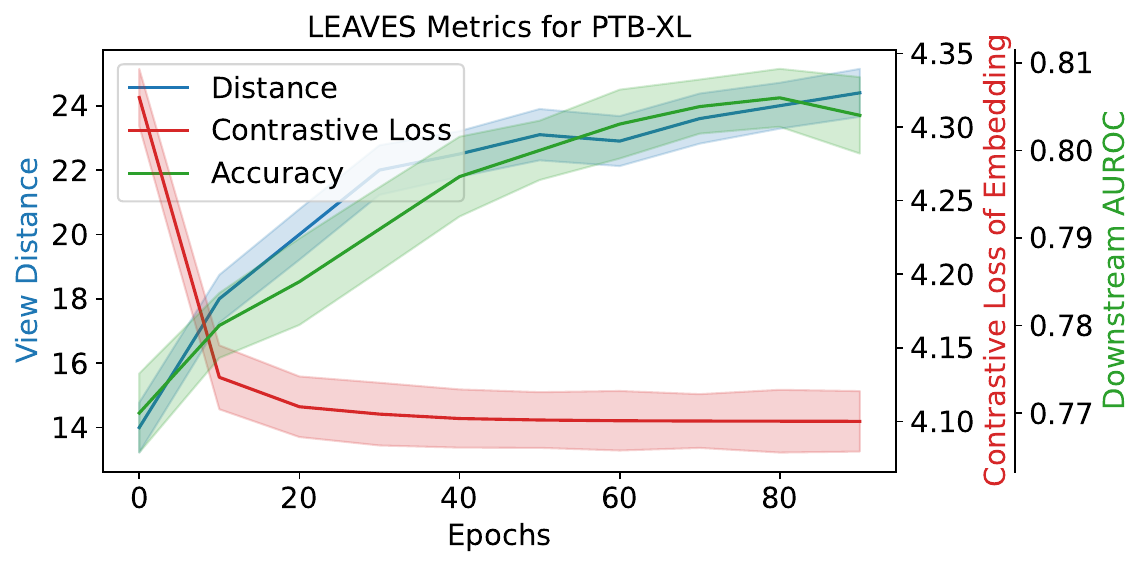}
  \vspace{-.5cm}
  \caption{Learning metrics of LEAVES (SimCLR) in datasets of PTB-XL. The blue lines represent the distance between the pairs of augmented signals calculated with dynamic time warping (DTW); red indicates the contrastive loss of the learning encoders; and green indicates the AUROC with frozen encoder.}
  \label{fig:leaves_metrics}
  \vspace{-.8cm}
\end{figure}

To further our understanding of the adaptive learning processes within the LEAVES framework, we analyze several key performance metrics throughout the training period. These metrics include the distance between paired views measured by dynamic time warping (DTW), contrastive loss, and AUROC in downstream tasks assessed using a linear probe with frozen encoders, as shown in Figure \ref{fig:leaves_metrics}.

The analysis reveals an increase in the distance between paired views, which suggests that the framework generates increasingly challenging views over time. Moreover, the improvement in AUROC on downstream tasks as training progresses demonstrates that the representations learned are increasingly effective. These results highlight the effectiveness of the LEAVES framework in achieving the dual objectives of contrastive learning: diversifying augmented views and enhancing encoder performance for varied applications.

\vspace{-0.3cm}
\subsection{Space \& Time Complexities}
The proposed method, LEAVES, has demonstrated significant advantages in both model space and time complexity when compared to previous state-of-the-art methods such as ViewMaker. Notably, LEAVES optimizes approximately 20 parameters for view generation, in contrast to the roughly 580K parameters required by the 1D ViewMaker structure. This substantial reduction in parameter count not only minimizes memory overhead but also contributes to faster convergence during training.

In addition, the integration of LEAVES into contrastive learning frameworks yields considerable runtime improvements. For instance, on the Sleep-EDFE dataset—using an \textit{AWS p3.2xlarge} instance equipped with an NVIDIA V100 GPU—the baseline SimCLR framework requires an average of 578.0 seconds per epoch (for 100 training epochs with a batch size of 128). When LEAVES is incorporated, the per-epoch time is reduced to 390.8 seconds. This efficiency gain is primarily attributed to the design of LEAVES, which integrates augmentations directly into the network. By doing so, it leverages GPU acceleration to perform complex augmentation operations concurrently with other network computations, thereby eliminating the need for separate data pre-processing and external augmentation modules that typically add computational overhead.


\subsection{Limitation}
Although we substantially improve the interpretability of the data augmentation process in contrastive learning by leveraging the pre-defined data transforming methods, challenges remain. The the optimization process is not easily understandable and makes it difficult for researchers to identify the best data augmentation policies. Therefore,  enhancing interpretability further would be beneficial to facilitate future research.



\section{Conclusion}
In this work, we introduce a simple but effective LEAVES module to learn augmentations for time-series data in contrastive learning. With an adversarial training approach, our proposed method optimizes the hyperparameters for data augmentation methods in contrastive learning. We evaluate the proposed method on four datasets, and find that it outperforms the baselines. 
Our evaluation across four diverse datasets demonstrates that LEAVES, as a computationally efficient solution, typically outperforms or closely matches existing baselines while avoiding extensive manual tuning. 
We also demonstrate that LEAVES can preserve the original information in augmented views, especially on ECG time-series data, compared to the prior methods. 

Although LEAVES achieves promising results in learning views for contrastive learning, there are still limitations. For example, the interpretability of the view-tuning process can be improved. Thus, future work will include enhancing interpretability to better understand data augmentation policies in contrastive learning. Also, we will expand more augmentation methods in LEAVES and apply LEAVES to a wider range of time-series data. 

\acks{This work was supported by National Science Foundation (\# 2047296 and \# 1840167) and National Institute of Health (\# R01DA059925)}

\bibliography{chil-sample}

\newpage
\appendix

\section{Additional Abalation Study}\label{apd:additional_ab}
\subsection{Ablation Study: Order of Augmentation Methods}
\label{ab:aug_order}
Does the order of augmentations matter? To answer this question, we conducted an ablation study to explore the effect of the order of the augmentations. For the four applications evaluated in Section \ref{evaluation}, we randomly shuffle the order of the augmentation methods applied and evaluate the performance in the downstream tasks. As control groups, we repeat the pre-training experiments with fixed augmentation orders but with shuffled random seeds. All ablation experiments are performed with the combination of LEAVES and SimCLR. Table \ref{tab:aug_order} shows the results of shuffling the order of augmentation methods. From the table, we can see that the performance variations of the shuffled order and the fixed order follow a similar pattern and that no statistically significant differences (unpaired t-test, p $>$ 0.05) are observed. Therefore, we can conclude that the order of the different augmentation methods applied in the LEAVES module does not significantly affect its performance.

\begin{table}[]
\centering
\caption{Performance impact of different augmentation method orders in LEAVES with SimCLR on downstream tasks. Macro-F1 score is used as the evaluation metric.}
\begin{tabular}{l|cc}
\hline
Dataset    & Random Order      & Fixed Order       \\ \hline \hline
Apnea-ECG  & 0.779 $\pm$ 0.011 & 0.781 $\pm$ 0.009 \\ 
Sleep-EDFE & 0.796 $\pm$ 0.014 & 0.791 $\pm$ 0.013 \\ 
PTB-XL     & 0.680 $\pm$ 0.010 & 0.678 $\pm$ 0.012 \\ 
PAMAP2     & 0.931 $\pm$ 0.009 & 0.934 $\pm$ 0.008 \\ \hline
\end{tabular}
\label{tab:aug_order}
\end{table}

\subsection{Ablation Study: Robustness Evaluation}
\begin{table*}[]
\centering
\caption{The macro-F1 performances table of applying pre-trained backbone on the clean and corrupted test sets. Diff. representations of the performance gap between clean and corrupted sets. The bold values indicate the smallest Diff. in performance between the clean and corrupted test sets.}
\begin{tabular}{ccccccc}
\hline
\multicolumn{1}{c|}{\multirow{2}{*}{Method}} & \multicolumn{3}{c|}{Apnea-ECG}                                                    & \multicolumn{3}{c}{Sleep-EDF}                                                    \\ \cline{2-7} 
\multicolumn{1}{c|}{}                        & Clean                & Corrupted            & \multicolumn{1}{c|}{Diff}           & Clean                & Corrupted            & \multicolumn{1}{c}{Diff}           \\ \hline \hline
\multicolumn{1}{c|}{Supervised}              & 0.757                & 0.671                & \multicolumn{1}{c|}{0.086}          & 0.769                & 0.688                & \multicolumn{1}{c}{0.081}          \\ 
\multicolumn{1}{c|}{SimCLR}                  & 0.775                & 0.705                & \multicolumn{1}{c|}{0.070}          & 0.783                & 0.729                & \multicolumn{1}{c}{0.054}          \\
\multicolumn{1}{c|}{BYOL}                    & 0.779                & 0.712                & \multicolumn{1}{c|}{0.067}          & 0.780                & 0.724                & \multicolumn{1}{c}{0.056}          \\ 
\multicolumn{1}{c|}{SimCLR(LEAVES)}         & 0.788                & 0.725                & \multicolumn{1}{c|}{0.063}          & 0.790                & 0.740                & \multicolumn{1}{c}{0.050}          \\
\multicolumn{1}{c|}{BYOL(LEAVES)}           & 0.795                & 0.733                & \multicolumn{1}{c|}{\textbf{0.062}} & 0.785                & 0.738                & \multicolumn{1}{c}{\textbf{0.047}} \\ \hline
\multicolumn{1}{l}{}                          & \multicolumn{1}{l}{} & \multicolumn{1}{l}{} & \multicolumn{1}{l}{}                & \multicolumn{1}{l}{} & \multicolumn{1}{l}{} & \multicolumn{1}{l}{}                \\ \hline
\multicolumn{1}{c|}{\multirow{2}{*}{Method}} & \multicolumn{3}{c|}{PAMAP2}                                                       & \multicolumn{3}{c}{PTB-XL}                                                       \\ \cline{2-7} 
\multicolumn{1}{c|}{}                        & Clean                & Corrupted            & \multicolumn{1}{c|}{Diff}           & Clean                & Corrupted            & \multicolumn{1}{c}{Diff}           \\ \hline \hline
\multicolumn{1}{c|}{Supervised}              & 0.896                & 0.812                & \multicolumn{1}{c|}{0.084}          & 0.662                & 0.615                & \multicolumn{1}{c}{0.047}          \\ \hline
\multicolumn{1}{c|}{SimCLR}                  & 0.925                & 0.867                & \multicolumn{1}{c|}{0.058}          & 0.671                & 0.619                & \multicolumn{1}{c}{0.052}          \\ \hline
\multicolumn{1}{c|}{BYOL}                    & 0.931                & 0.868                & \multicolumn{1}{c|}{0.063}          & 0.671                & 0.621                & \multicolumn{1}{c}{0.050}          \\ \hline
\multicolumn{1}{c|}{SimCLR(LEAVES)}         & 0.934                & 0.883                & \multicolumn{1}{c|}{\textbf{0.051}} & 0.669                & 0.632                & \multicolumn{1}{c}{\textbf{0.037}} \\ \hline
\multicolumn{1}{c|}{BYOL(LEAVES)}           & 0.936                & 0.881                & \multicolumn{1}{c|}{0.055}          & 0.677                & 0.636                & \multicolumn{1}{c}{0.041}          \\ \hline
\end{tabular}
\label{tab:robustness}
\end{table*}

We further conduct an ablation study to examine the robustness of augmentation-based contrastive learning methods under noise corruption. In this experiment, we perform supervised learning on top of the pre-trained backbones using the original and LEAVES-integrated SimCLR and BYOL. Differing from the evaluation setting in Section~\ref{evaluation}, we fine-tune the models with clean training sets. However, we evaluate the model performance on the corrupted test set with randomly generated noises from \textit{Jittering, Scaling, MagW}, and \textit{TimeM} with randomized $\sigma$ between 0.01 and 0.05. The corruption applied to the test set thus introduces distribution shifts between the original and augmented datasets. Table~\ref{tab:robustness} shows the results of the robustness evaluation. We conduct paired t-tests to compare the \textit{Diff} between clean and corrupted samples in terms of robustness. We compare the results from 4 datasets of supervised learning to those from SimCLR and BYOL, respectively. Additionally, we conduct paired t-tests between SimCLR and BYOL, and between SimCLR (LEAVES) and BYOL (LEAVES). The test results show that all the p-values were below 0.05. However, when we compare SimCLR to BYOL and SimCLR (LEAVES) to BYOL (LEAVES), no significance can be observed (p $>$ 0.05). From the table and the results of statistical tests, we observe that applying both SimCLR and BYOL pre-training methods improved the model's robustness to corrupted noises. Moreover, by integrating the LEAVES module, the negative impact of introducing noise into the test set is further alleviated. On the other hand, the choice between SimCLR and BYOL does not affect the improvement in model robustness.

\section{Implementation Details}

\subsection{Hyperparameter and Training Details}

All models were implemented in PyTorch and trained on an NVIDIA V100 GPU. For both the self-supervised pre-training and the downstream supervised fine-tuning, we used the Adam optimizer. The learning rate was set to \texttt{1e-3} for pre-training and fine-tuned for each downstream task. A consistent batch size of 128 was used across all experiments. The temperature parameter ($\tau$) for the SimCLR contrastive loss was set to 0.05.

The pre-training phase for all contrastive models was run for 100 epochs. For the downstream fine-tuning, the number of epochs varied by dataset to ensure convergence, as detailed in Table 5.

\begin{table}[h]
\centering
\begin{tabular}{|l|c|c|c|c|}
\hline
\textbf{Hyperparameter} & \textbf{Apnea-ECG} & \textbf{Sleep-EDF} & \textbf{EPTB-XL} & \textbf{PAMAP2} \\
\hline
\multicolumn{5}{|c|}{\textbf{Pre-training}} \\
\hline
Optimizer & Adam & Adam & Adam & Adam \\
Learning Rate & 1e-3 & 1e-3 & 1e-3 & 1e-3 \\
Batch Size & 128 & 128 & 128 & 128 \\
Epochs & 100 & 100 & 100 & 100 \\
\hline
\multicolumn{5}{|c|}{\textbf{Fine-tuning}} \\
\hline
Optimizer & Adam & Adam & Adam & Adam \\
Learning Rate & 1e-4 & 1e-4 & 5e-5 & 5e-5 \\
Batch Size & 128 & 128 & 128 & 128 \\
Epochs & 50 & 50 & 80 & 80 \\
\hline
\end{tabular}
\caption{Hyperparameter settings for pre-training and downstream fine-tuning across all datasets.}
\label{tab:hyperparameters}
\end{table}

\subsection{Downstream Task Evaluation: Full Finetuning}

To clarify the evaluation process for the main results presented in Table 1, we employed a full fine-tuning protocol. After the self-supervised pre-training phase, the learned encoder weights were used to initialize a 1D ResNet-18 model. A linear classification head was then added on top of the encoder. Subsequently, the \textbf{entire network} (both the encoder and the new classifier head) was fine-tuned end-to-end on the labeled training data of each respective downstream task. This approach allows the model to adapt the learned representations specifically to the target task, which typically leads to superior performance compared to using a frozen encoder with a linear probe. The Supervised baseline was trained from a random initialization using the identical architecture and fine-tuning procedure for a fair comparison.

\subsection{Code Availability}

The source code and trained models for this research are publicly available on GitHub to ensure reproducibility and facilitate future research: \url{https://github.com/comp-well-org/LEAVES}.

\end{document}